\documentclass[10pt,twocolumn,letterpaper]{article}
\usepackage[pagenumbers]{cvpr} % To force page numbers, e.g. for an arXiv version

\usepackage{graphicx}
\usepackage{amsmath}
\usepackage{amssymb}
\usepackage{booktabs}
\usepackage[colorlinks=true, allcolors=blue]{hyperref}
\usepackage{comment}
\usepackage{afterpage}

\makeatletter
\newenvironment{tablehere}
{\def\@captype{table}}
{
	
}

\makeatother

\usepackage[capitalize]{cleveref}
\crefname{section}{Sec.}{Secs.}
\Crefname{section}{Section}{Sections}
\Crefname{table}{Table}{Tables}
\crefname{table}{Tab.}{Tabs.}

\def\cvprPaperID{*****} % *** Enter the CVPR Paper ID here
\def\confName{CVPR}
\def\confYear{2022}

\usepackage{array}
\usepackage{url}

\newcommand\Mark[1]{\textsuperscript#1}

\pdfoutput=1

\begin{document}

\title{ Identification of Cognitive Workload during Surgical Tasks with Multimodal Deep Learning }

% \author{Kaizhe Jin, Adrian Rubio-Solis, Ravi Naik, Tochukwu Onyeogulu, Amirul Islam, Salman Khan \\ Izzeddin Teeti, James Kinross} 
% %  Daniel R Leff\Mark{1}\Mark{,}\Mark{2},  Fabio Cuzzolin\Mark{3} and George Mylonas\Mark{1}\Mark{,}\Mark{2}
\author{Kaizhe Jin\Mark{1}\Mark{,}\Mark{2}, Adrian Rubio-Solis\Mark{1}\Mark{,}\Mark{2}, Ravi Naik\Mark{1}\Mark{,}\Mark{2}, Tochukwu Onyeogulu\Mark{3}, Amirul Islam\Mark{3}, Salman Khan\Mark{3} \\ Izzeddin Teeti\Mark{3}, James Kinross\Mark{2}, 
 Daniel R Leff\Mark{1}\Mark{,}\Mark{2},  Fabio Cuzzolin\Mark{3} and George Mylonas\Mark{1}\Mark{,}\Mark{2}
\\
{\tt\small  k.jin20, arubioso, ravi.naik15,  j.kinross, d.leff, george.mylonas@imperial.ac.uk}
\\
{\tt\small 19175136, aislam, salmankhan, iteeti, fabio.cuzzolin@brookes.ac.uk}
\\
\Mark{1} Hamlyn Centre for Robotic Surgery, Imperial College London, London SW7 2AZ, UK
\\
\Mark{2} Department of Surgery $\&$ Cancer, Imperial College London, London SW7 2AZ, UK
\\
\Mark{3} Visual Artificial Intelligence Laboratory, Oxford Brookes University, Oxford, OX33 1HX, UK
%++++++++++++++++++++++++++++++++++++++++++++++++++++
% \\
% \Mark{3}Department1
% \and
}
\maketitle

%%%%%%%%% ABSTRACT
\begin{abstract}
The operating room (OR) is a dynamic and complex environment consisting of a multidisciplinary team working together in a high take environment to provide safe and efficient patient care. Additionally, surgeons are frequently exposed to multiple psycho-organisational stressors that may cause negative repercussions on their immediate technical performance and long-term health. Many factors can therefore contribute to increasing the Cognitive Workload (CWL) such as temporal pressures, unfamiliar anatomy or distractions in the OR.  In this paper, a cascade of two machine learning approaches is suggested for the multimodal recognition of CWL in four different surgical task conditions. Firstly, a model based on the concept of transfer learning is used to identify if a surgeon is experiencing any CWL. Secondly, a Convolutional Neural Network (CNN) uses this information to identify different degrees of CWL associated to each surgical task. The suggested multimodal approach considers adjacent signals from electroencephalogram (EEG), functional near-infrared spectroscopy (fNIRS) and eye pupil diameter. The concatenation of signals allows complex correlations in terms of time (temporal) and channel location (spatial). Data collection was performed by a Multi-sensing AI Environment for Surgical Task $\&$ Role Optimisation platform  (MAESTRO) developed at the Hamlyn Centre, Imperial College London. To compare the performance of the proposed methodology, a number of state-of-art machine learning techniques have been implemented. The tests show that the proposed model has a precision of 93$\%$.

\end{abstract}

%%%%%%%%% BODY TEXT
\section{Introduction}
\label{sec:intro}
The Operating Room  (OR) can be described as a complex socio-technical environment where activities are influenced by the environment, team dynamics and patient physiology.  Working in the OR exposes the clinical team to multiple psycho-organizational challenges that can results in negative repercussions on their health and on their performance at work \cite{sami2012real, uugurlu2015effects}. As pointed out in \cite{dias2018systematic}, surgeons and clinicians in the OR are usually exposed to factors that might increase the associated cognitive workload (CWL) as a consequence of dealing with high-demand tasks that require simultaneous processing of large amounts of information potentially leading to cognitive overload and performance decline.

Regardless of how competent a surgeon may be, they are not immune to their own cognitive limitations, frailties, and fallibilities that characterise the human brain \cite{dias2018systematic, ortega2021deep}. CWL is a term used to describe the phenomenon of working memory use. In the OR, stressors or factors that can lead to increased CWL include team communication, noise, simultaneous processing of visual and auditory information as well as surgeon-related factors such stress and fatigue \cite{dias2018systematic}. Although the sensory memory system is able to process large amounts of visual and auditory information, working memory in humans has a limited capacity to process this information simultaneously. The allocation of limited working memory to a particular task is therefore crucial when undertaking high-risk technical tasks. 

It is generally accepted that surgeons with greater experience are capable of managing higher levels of CWL compared to novices, particularly when balancing operative task demands against the available cognitive resources \cite{chiarelli2018deep, tuauctan2021dimensionality}. However, on occasions surgeons and clinicians experience a state of cognitive overload, in particular in those that involve complex and/or non-routine operative situations such as emergencies and unexpected events.

Presently, a number of research efforts have been produced in the concatenation of combined physiological signals to quantify human mental workload \cite{tuauctan2021dimensionality, li2016ecg, guerrero2021eeg}. For example in \cite{guerrero2021eeg}, a hybrid functional neuroimaging technique using electroencephalography (EEG) and functional near-infrared spectroscopy (fNIRS) as imaging modalities with 17 healthy subjects performing the letter n-back task was suggested. Such text is a standard experimental paradigm related to the human capabilities of working memory (WM). 
\begin{figure}[!t]
\centering
\includegraphics[width= 0.54 \textwidth]{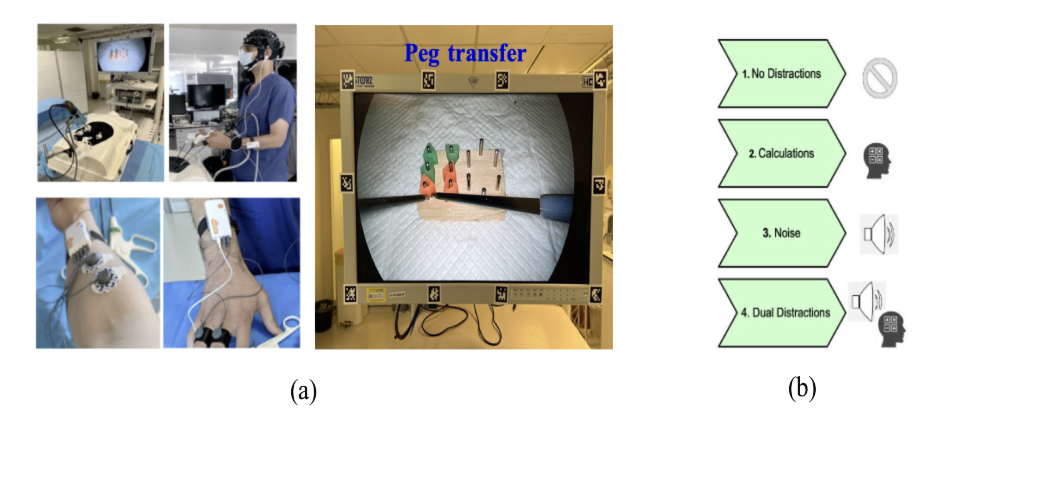}
\caption{\label{fig:Peg_transfer} (a) Surgical Setup and wearable sensors, and (b) Laparoscopic Peg Transfer under four different conditions in randomised order.}
\end{figure}

In \cite{guerrero2021eeg}, during each experiment, the mental workload associated with each task was parametrically updated, in which nineteen EEG channels covering the whole head and 19 fNIRS channels were located on the forehead to cover the most dominant brain region involved in WM. In [8], it was demonstrated that robust classification of brain signals can be achieved by combining EEG signals with other neuroimaging signals such as with fNIRS. Unlike conventional machine learning techniques where an a priori feature extraction step is necessary to train a model, with deep learning approaches signals can be fed directly. In \cite{guerrero2021eeg}, results showed that not only classification accuracy was significantly improved, but also additional data processing was eliminated.
\begin{figure}[!t]
\centering
\includegraphics[width= 0.58 \textwidth]{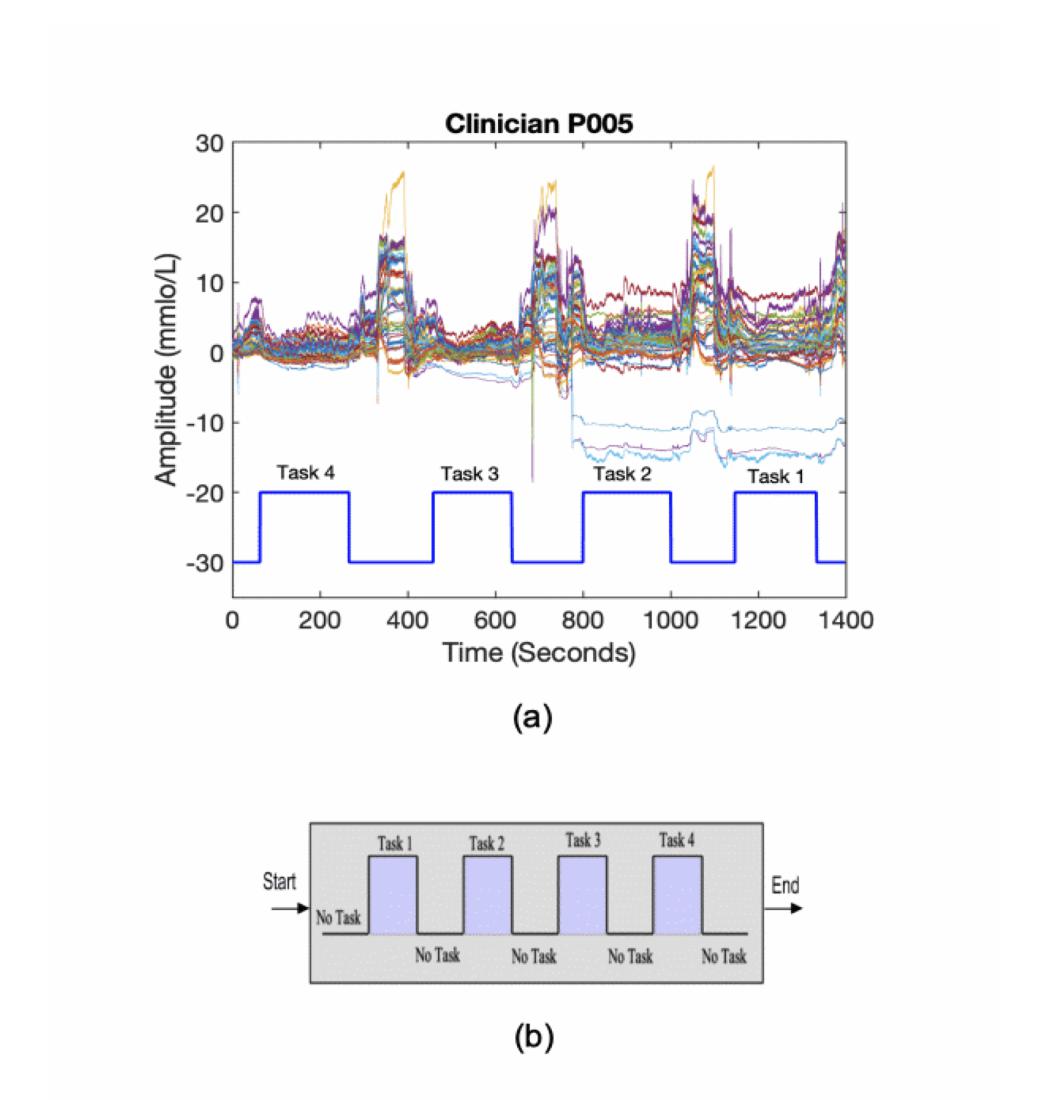}
\caption{\label{fig:boxcar} (a) fNIRS signals that corresponds to Subject five (P005) used as an example and (b) Boxcar of tasks performed in a random experiment and performed by each surgical resident. }
\end{figure}
In this research work, a systematic approach based on a Multimodal Deep learning strategy is suggested for the identification of CWL experience by a group of five interns in a simulated OR. Surgical operations involve a pilot study related to a Laparoscopic peg transfer (LPT) task as illustrated in figure \ref{fig:Peg_transfer}.

In each experiment, LPT is undertaken under four different conditions and designed in a way to increase the mental demand on surgeon. This is achieved by increasing introducing a neurocognitive task or auditory distraction. As described in \cite{singh2018impact}, the impact of increasing distracting stressors may contribute to an increase in the associated cognitive workload (CWL).  In particular, novice surgeons may sacrifice attention and decision making on one task in order to level out their ability on another \cite{modi2017decade}. Opposite to this, experienced surgeons usually posses a greater cognitive capacity to perform a surgical task under additional distractors and mental load without a significant decrease in their performance \cite{singh2018impact}. As suggested in  \cite{modi2017decade}, the recruitment of the prefrontal cortex scales linearly to cope with an increase in the cognitive demand. To test this hypothesis, the impact of additional distraction stressors on the  during surgical tasks is evaluated based on the average HbO$_2$ activation response of interns.

In this work, the suggested multimodal deep learning is applied to identify a surgical task having different distractions that may contribute to an increase of the CWL. Therefore, additional distractors for each experiment involve ( See figure \ref{fig:Peg_transfer}):
\begin{figure*}[!t]
\centering
\includegraphics[width= 0.7 \textwidth]{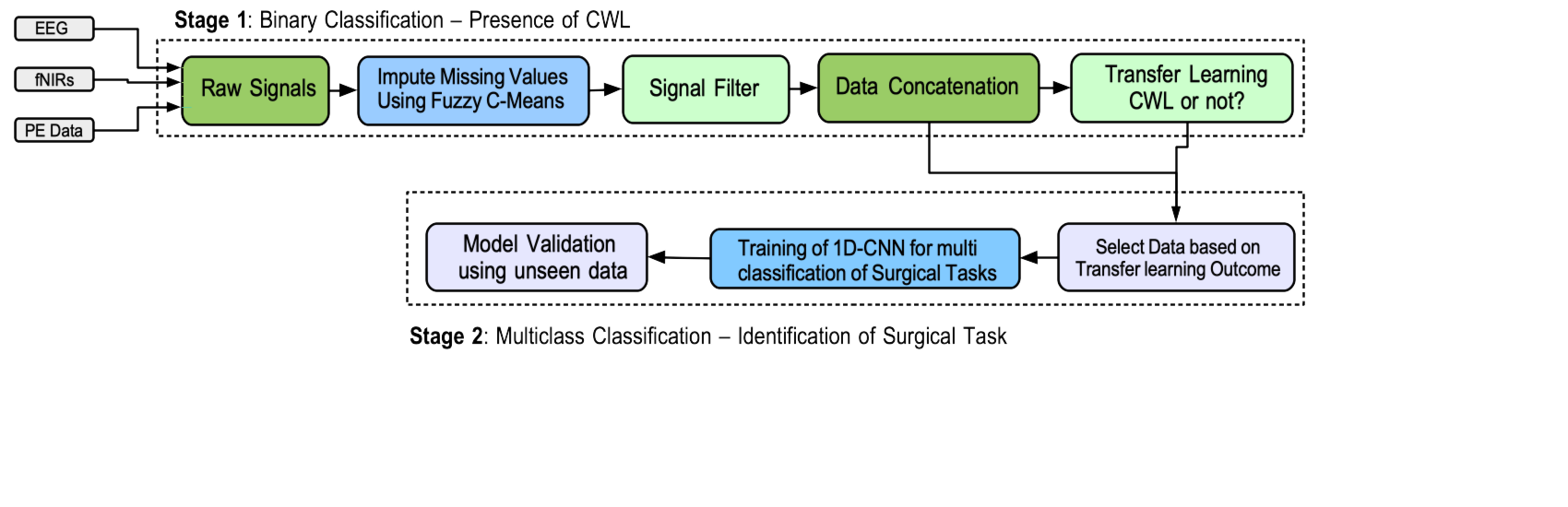}
\caption{\label{fig:flow_diagram}(a) Flow diagram for signal processing and Multimodal Deep Learning Strategy for the classification of surgical tasks using concatenated EEG, fNIRs and Pupil Eye Data.}
\end{figure*}
\begin{itemize}
    \item Task 1: Laparoscopic Peg Transfer (LPT) only. 
\item Task 2: LPT + Serial subtractions of 7 from 1000.
\item Task 3: LPT +  Recurring hospital bleep sound at 80 dB.
\item Task 4: Dual distractions with tasks 3 and 4 undertaken simultaneously. 
\end{itemize}
Once each task is completed, each intern assumes a motor rest position in which no action is performed and that we termed the 'No Task' for short where there was no additional CWL applied from either the task or distraction. 

The proposed multimodal approach, is a two-step strategy in which, concatenated EEG, fNIRS and pupil diameter of both eyes of each intern is used as input data. In the first step, the concept of Transfer Learning (TL) using an AlexNet network is applied to identified when an intern is experiencing an increase the CWL. In the second step, the type of surgical task is identified and the concentration of HbO$_2$ is evaluated to estimate the associated cognitive demand.

The rest of this paper is divided as follows. In section 2, data collection and the proposed methodology is described. Experiments and results are provided in section 3, while future work and conclusions are drawn in section 4. 
% %++++++++++++++++++++++++++++++++++++++++
\section{Methods}
\subsection{Data Collection}
Five healthy subjects were recruited and informed consent was obtained. Each particpant was fitted with the relevant sensors and asked to complete the LPT under each on the four conditions in a randomised order. Each task started wtih a baseline measurement of physiological signals (No Task) and were asked to complete the SURG-TLX questionnaire to obtain subjective results on perceived workload at the end of each task condition. The time to complete each experiment is approximately 29 minutes. In this work, data from three different sensing sources are used to train the proposed model. First, fNIRS data were collected using the Artinis 24 $\times$ 11 system, (Brite, Artinis Medical Systems, Elst, The Netherlands, www.artinis.com), across 22 prefontal cortex locations. In Figure 2(a), sample raw fNIRS data that correspond to subject P005 are illustrated, and their corresponding boxcar of events randomly selected. Secondly, EEG signals were collected from a TMSi Mobita  32 channel system (TMSi, Mobita, EEG system , The Nederlands).  Thirdly, data corresponding to the left and right eye pupil diameter were collected using eye tracker glasses  (Pupil Labs Pupil Core).

% \begin{figure*}[t!]
% \begin{center}
% \includegraphics[width=16.1cm, height=10.0cm ]{Figures/AlexNet.png }
% \caption{\footnotesize (a) Convolutional Neural Network (AlexNet) used for Transfer Learning in the detection of any presence of CWL and (b) 1D CNN applied to multiclass Classification for the identification of the type of CWL in each task.
% .}\label{fig::ROV_components}
% \end{center}
% \end{figure*}
The CWL of the intern performing Task 1 is expected to be much lower than that of tasks 2, 3 and 4. We define three categories for the CWL as follows:

\begin{itemize}
    \item Task 1 - Normal CWL (CWL-1 for short)
    \item Task 2 - CWL of type 1 (CWL-2 for short)
    \item Task 3 - CWL of type 2 (CWL-3 for short)
    \item Task 4 - High CWL (CWL-4 for short)
\end{itemize}

Note: For each surgical experiment, the order used for tasks 1-4 are randomly suggested. From figure \ref{fig:boxcar},  the execution of a task is represented by a step, while for no task a zero signal is used.
\begin{figure*}[!t]
\centering
\includegraphics[width= 0.7 \textwidth]{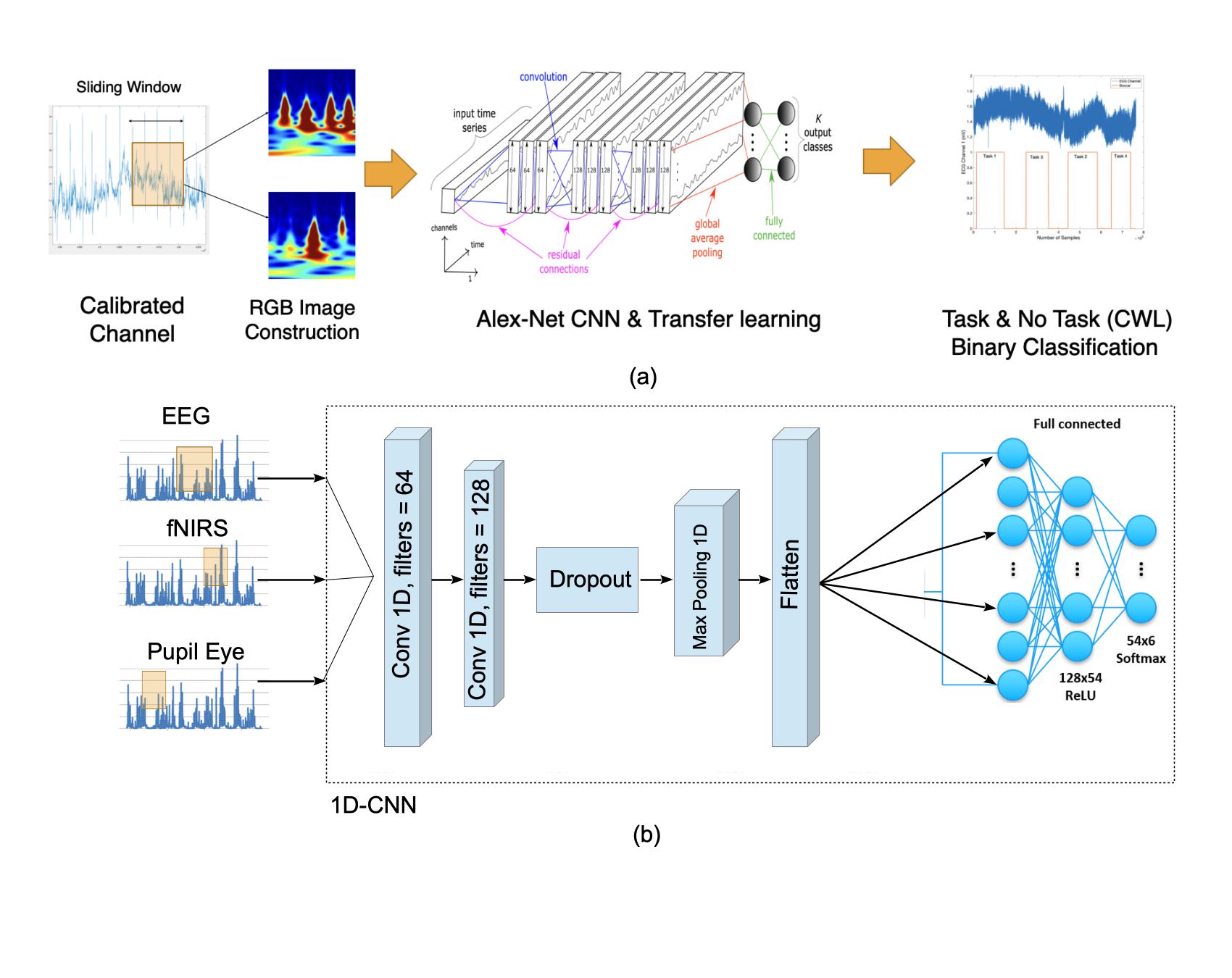}
\caption{\label{fig:Network_AlexNet_1DCNN}(a) Convolutional Neural Network (AlexNet) used for Transfer Learning in the detection of any presence of CWL and (b) 1D CNN applied to multiclass Classification for the identification of the type of CWL in each task.}
\end{figure*}
% %++++++++++++++++++++++++++++++++++++++++++++++++++
\subsection{Proposed Multimodal Deep Learning Strategy}
To identify each task,  a multimodal strategy that consists of two sequential deep learning approaches based on the concept of Transfer Learning and Convolutional Neural Networks (CNNs) has been suggested. As illustrated in figure \ref{fig:flow_diagram}, in the first step, raw data is collected at a frequency of 1000Hz for EEG and fNIRs and 120Hz for pupil eye data (PE) respectively. For data modelling, sampling rate for EEG and fNIRs has been downsampled to 500Hz. Due to the eye blinking, missing data are intrinsically created for the collection of pupil diameter of each eye. To impute each missing value, the data clustering approach developed in \cite{hathaway2001fuzzy} has been applied. 

As described in figure \ref{fig:flow_diagram}, a data filter is implemented for each signal using a 5th order high-pass Butterworth filter with a cut-off frequency at 0.5 Hz. The cleaned concatenated signal was segmented into a 400-sample sliding window. After segmentation, three channels were concatenated horizontally, forming a (848, 1) feature vector.

In the first step, an AlexNet network is applied to identify the presence of any CWL during each experiment. This  classification problem is viewed as  a binary prediction about the presence of any type of increase in the CWL that may experience each surgeon by performing a given surgical task. In transfer learning, the AlexNet is a convolutional neural network (CNN) that has been successfully applied to a number of classification problems where only its last layer is trained \cite{alom2018history}.

The AlexNet network has been trained on over a million images and can classify images into 1000 object categories such as animals, objects, keyboard, etc. The AlexNet network has been trained to recognise a  number of rich feature representations for a wide range of images \cite{alom2018history}. The network takes an image as input and outputs a label for the object in the image together with the probabilities for each of the object categories. Transfer learning theory has been frequently used in deep learning applications as a pre-trained network and use it as a starting point to learn a new task for image classification \cite{aghajani2017measuring,saadati2019multimodal, zakeri2020physiological,lu2019pathological }.

In Transfer Learning, fine-tuning is usually a much faster and easier learning process than training a network with randomly initialized weights from scratch. Previous learned knowledge and features can be easily transferred to a new task using a smaller number of training images. In this work, EEG, fNIRS and PE data have been concatenated in a 1D vector array and then their associated scalogram image has been obtained using Continuous Wavelength Transform (CWT) \cite{yildirim2018arrhythmia }. CWT produces a scalogram plot along with its corresponding time-based frequency spectrum plot. Using CWT a set of high level features of the dominant frequencies and the related scales are extracted and can be used to train and validate a signal classifier based on the model of neural networks.  

To recapitulate, a binary classification of surgical activities is initially performed to detect if there is any atypical CWL due to the presence of distractors or not. If any atypical increase in CWL is detected, then a second stage in the proposed multimodal strategy is triggered and used to recognise the type of CWL solved as a classification problem (multiclass classification as illustrated in figure \ref{fig:flow_diagram}). From figure \ref{fig:Network_AlexNet_1DCNN}(a), an initial model based on the popular deep learning neural network AlexNet is used as the baseline model to identify the presence of atypical CWL. In the second step, a 1D convolutional neural network is implemented, in which the 1D vector array that results from concatenating EEG, fNIRS and PE is used as input training data. The 1D CNN used in this work follows the neural structure presented in figure \ref{fig:Network_AlexNet_1DCNN}(b). Finally, unseen data is used to validate the proposed multimodal deep learning strategy.  

\section{Experiments and Results}
In the following sections the results that correspond to the binary (detection of a level of CWL or not due to the presence of surgical tasks) and multiclass classification (recognition of task 1-4) are presented. 
\subsection{Binary Classification: Detection of CWL}
In this section, the binary classification results obtained by the transfer learning approach are presented. As described above, the binary classification approach suggested in this work aims at determining the presence of a surgical task or not which we assume is directly related to the presence of CWL or not. We define two main classes, i.e. a) Clinician task - CWL and b) No Clinician Task. 

A data set that involves 22 channels for fNIRS and 18 for EEG and 2 channels for pupil diameter (PE) respectively, and a number of 200 samples in a sampling window was found to be optimal. The average of all EEG signals and average of all channels for fNIRS was calculated and then both were concatenated in an input matrix $X_i$= 200 samples. Dataset was split into two subsets, i.e. a) 70$\%$ for training, and b) 30 $\%$ for validation the suggested model based on transfer learning. 

\begin{figure}[!t]
\centering
\includegraphics[width= 0.47 \textwidth]{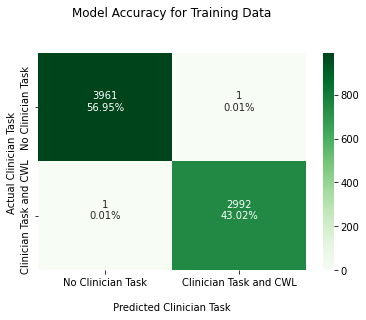}
\includegraphics[width= 0.47 \textwidth]{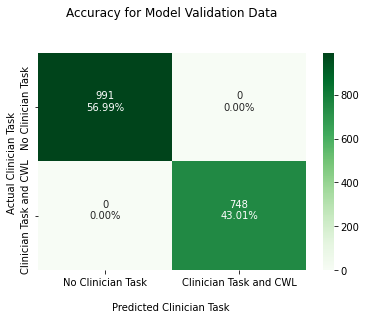}
\caption{\label{fig:binary_confusion} Average model accuracy using a confusion matrix for the identification of any presence of CWL during each experiment for training and testing}
\end{figure}

In Table \ref{table:binary_classification}, a number of traditional machine learning algorithms have been implemented and used to compared the average performance of the proposed multimodal approach. Such techniques include Random Forest and Support Vector Machines (RF + SVM), Random Forest and Principal Component Analysis (RF + PCA), Transfer Learning (TF), 1D convolutional neural network, Multilayer Extreme Learning Machine (MLELM), and Extreme Learning Machine applied to a Perceptron neural network. Both RF + SVM and RF + PCA employs RF as a feature extraction method to achieve a high level representation of input data. 

From Table \ref{table:binary_classification}, it can also be observed that the highest trade-off between model accuracy and training time is produced by the model MELM. However, for online implementation and validation of new data, 1D-CNN and TL offers the highest performance, in which all the three signals, i.e. EEG, fNIRS and PE are used as input data.
\vspace{5mm}
%++++++++++++++++++++++++
\begin{tablehere}
\captionsetup{font=footnotesize, justification=centering}
\caption{ Average binary classification of different machine learning algorithms and the proposed multimodal deep learning approach.}\label{iris_results} % title of Table
%\vspace{0.0cm}
\centering % used for centering table
\begin{tabular}{p{1.50cm} |p{0.1cm}|p{0.1cm} |p{0.9cm}|p{0.1cm} |p{0.1cm}| p{0.1cm} | p{1.5cm} }
\hline 
%inserts double horizontal lines
 %& \multicolumn{2}{c}{RMSE} \\
 \scriptsize Model &\multicolumn{3}{c}{\scriptsize Training ($\%$)}  & \multicolumn{3}{c}{\scriptsize Testing ($\%$)} & \scriptsize Sensors\\
%[-0.4em] 
%......................................................
\hline
& \multicolumn{2}{c}{\scriptsize Average}&\scriptsize Time (s)&\multicolumn{3}{c}{\scriptsize Average}&\multicolumn{1}{c}{ \scriptsize  }\\   
\hline
&  \multicolumn{7}{c}{ \scriptsize Binary Classification }\\   
\hline 
%...................................................... 
% \multicolumn{8}{c}{\scriptsize  \textbf{MNIST} } \\
% \hline
  \scriptsize  RF + SVM & \multicolumn{2}{c}{\scriptsize 85.1 } & \multicolumn{1}{c}{\scriptsize 2490 } &\multicolumn{3}{c}{\scriptsize 70.2 } & \multicolumn{1}{c}{\scriptsize fNIRS + EEG}\\
 \scriptsize  RF + PCA& \multicolumn{2}{c}{\scriptsize 88.2 } & \multicolumn{1}{c}{\scriptsize  2400} &\multicolumn{3}{c}{\scriptsize 73.4} & \multicolumn{1}{c}{\scriptsize  fNIRS + EEG }\\
  \scriptsize  1D-CNN & \multicolumn{2}{c}{\scriptsize 100} & \multicolumn{1}{c}{\scriptsize 318 } &\multicolumn{3}{c}{\scriptsize 99.8}&\multicolumn{1}{c}{\scriptsize  fNIRS + EEG + PE}\\
  \scriptsize TL  & \multicolumn{2}{c}{\scriptsize 100 } & \multicolumn{1}{c}{\scriptsize 980} &\multicolumn{3}{c}{\scriptsize  99.9}&\multicolumn{1}{c}{\scriptsize  fNIRS + EEG + PE }\\
  \scriptsize MELM  & \multicolumn{2}{c}{\scriptsize 98.7} & \multicolumn{1}{c}{\scriptsize 78.4} &\multicolumn{3}{c}{\scriptsize 94.5}&\multicolumn{1}{c}{\scriptsize  fNIRS + EEG + PE}\\
  \scriptsize ELM  & \multicolumn{2}{c}{\scriptsize 71.9} & \multicolumn{1}{c}{\scriptsize 40.5} &\multicolumn{3}{c}{\scriptsize 60.0}&\multicolumn{1}{c}{\scriptsize  fNIRS + EEG + PE }\\   
\hline
%====================================================
%**************************************************
	%	[1ex] % [1ex] adds vertical space
		
		\end{tabular}
		\centering % used for centering table
		\label{table:binary_classification} % is used to refer this table in the text
\end{tablehere}

 Finally, in figure \ref{fig:binary_confusion}, the confusion matrix that corresponds to the average cross-validation (training and validation) of five random experiments is presented.

\begin{figure}[!t]
\centering
\includegraphics[width= 0.47 \textwidth]{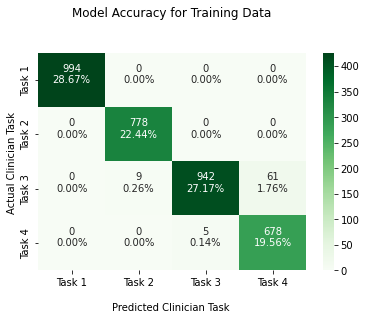}
\includegraphics[width= 0.47 \textwidth]{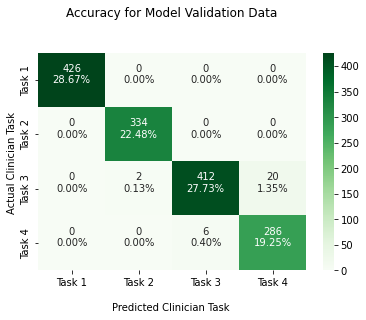}
\caption{\label{fig:muticlass_confusion} Average multiclass classification of the proposed multimodal deep learning approach for training and validation data from using EEG, fNIRS and PE data. }
\end{figure}
% \begin{figure}
% \begin{center}
% %\centering
% %\vspace{2mm}
% \includegraphics[width=6.5cm, height=6.2cm ]{Figures/training_multiclass.png }
% \includegraphics[width=6.5cm, height=6.2cm ]{Figures/testing_multiclass.png }
% \caption{\footnotesize Average multiclass classification of the proposed multimodal deep learning approach for training and validation data from using EEG, fNIRS and PE data.}\label{fig::Piscina}
% \end{center}
% \end{figure}

\subsection{Multiclass Classification of Surgical Tasks}

In this section, the results that correspond to the recognition of the different types of CWL that may results from each task 1-4 are presented. A data set of  3392 samples were constructed from raw fNIRS + EEG and PE data are used to train the proposed 1D Convolutional Neural Network (CNN) for the classification of task 1-4.  

In figure \ref{fig:muticlass_confusion}, the average accuracy of five experiments for the training and validation of the proposed multimodal deep learning (second stage: 1D CNN) strategy is presented, which achieved a model accuracy of 97$\%$. From figure \ref{fig:muticlass_confusion}(a), it is clear than an overall accuracy of more than 95$\%$ for the training of the proposed model is achieved. Similarly, for its validation, it can be noted that for tasks 2 and 3, an average number of 3 misclassifications is produced. 
% As can bFrom Figure 7(a), it can be noted  that the misclassification of tasks during the training of a 1D CNN is smaller than 2%. Similarly, from Figure 7(b), the suggested model produces a low classification error for the prediction (we use this term for classifying unseen data) of unseen data (validation or testing) even smaller than 2%. 

From both confusion matrices, it is clear the proposed neural model provides a good trade-off between model simplicity in terms of deep learning structures and a high testing accuracy while a number of only 200 samples is required to provide an average accuracy of about 98$\%$ as illustrated in figure \ref{fig:muticlass_confusion}(b). Moreover, the accuracy for validation shows that the proposed multimodal approach is capable in correlating  temporal and spatial information from brain signals and physiological signals. 

\begin{figure}[!t]
\centering
\includegraphics[width= 0.47 \textwidth]{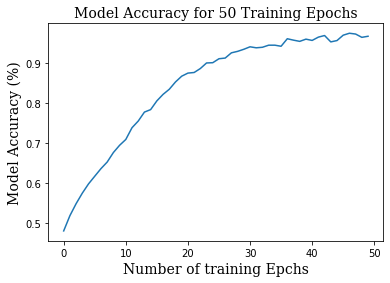}
\caption{\label{fig:evolution } Evolution of the model accuracy for binary classification}
\end{figure}

In figure \ref{fig:evolution }, the evolution of the training accuracy for the binary and multi class classification for 50 and 80 epochs is illustrated. In Table \ref{table:nfm_results}, the classification accuracy for tasks 1-4 is presented. According to our results, in terms of model simplicity vs model training, MELM is significantly superior to other deep learning algorithms. 

% \begin{figure}
% \begin{center}
% %\centering
% %\vspace{2mm}
% \includegraphics[width=5cm, height=5.3cm ]{Figures/5_classes_ECG_Accuracy_Hist.png }
% \includegraphics[width=5cm, height=5.3cm ]{Figures/hist_of_accuracy.png }
% \caption{\footnotesize Evolution of the training accuracy for the TL and 1D-CNN models respectively.}\label{fig::Piscina}
% \end{center}
% \end{figure}

\begin{table}
\captionsetup{font=footnotesize, justification=centering}
\caption{ Average performance for five-cross-validation and on-board classification.}\label{iris_results} % title of Table
%\vspace{0.0cm}
\centering % used for centering table
\begin{tabular}{p{1.70cm} |p{0.1cm}|p{0.1cm} |p{0.9cm}|p{0.1cm} |p{0.1cm}| p{0.1cm} | p{1.5cm} }
\hline 
%inserts double horizontal lines
 %& \multicolumn{2}{c}{RMSE} \\
 \scriptsize Model &\multicolumn{3}{c}{\scriptsize Training ($\%$)}  & \multicolumn{3}{c}{\scriptsize Testing ($\%$)} & \scriptsize Sensors\\
%[-0.4em] 
%......................................................
\hline
& \multicolumn{2}{c}{\scriptsize Average}&\scriptsize Time (s)&\multicolumn{3}{c}{\scriptsize Average}&\multicolumn{1}{c}{ \scriptsize  }\\   
\hline
&  \multicolumn{7}{c}{ \scriptsize Multiclass Classification }\\   
\hline 
%...................................................... 
% \multicolumn{8}{c}{\scriptsize  \textbf{MNIST} } \\
% \hline
  \scriptsize  RF + SVM & \multicolumn{2}{c}{\scriptsize 80.1 } & \multicolumn{1}{c}{\scriptsize 2390 } &\multicolumn{3}{c}{\scriptsize 50.2 } & \multicolumn{1}{c}{\scriptsize fNIRS + EEG}\\
 \scriptsize  RF + PCA& \multicolumn{2}{c}{\scriptsize 82.3 } & \multicolumn{1}{c}{\scriptsize  2400 } &\multicolumn{3}{c}{\scriptsize 54.4 } & \multicolumn{1}{c}{\scriptsize  fNIRS + EEG }\\
  \scriptsize  1D-CNN & \multicolumn{2}{c}{\scriptsize 97.5 } & \multicolumn{1}{c}{\scriptsize 318 } &\multicolumn{3}{c}{\scriptsize 93.7}&\multicolumn{1}{c}{\scriptsize  fNIRS + EEG + PE}\\
  \scriptsize TL  & \multicolumn{2}{c}{\scriptsize 98.4 } & \multicolumn{1}{c}{\scriptsize 5219} &\multicolumn{3}{c}{\scriptsize  92.6}&\multicolumn{1}{c}{\scriptsize  fNIRS + EEG + PE }\\
  \scriptsize MELM  & \multicolumn{2}{c}{\scriptsize 99.3} & \multicolumn{1}{c}{\scriptsize 200.2} &\multicolumn{3}{c}{\scriptsize 90.8}&\multicolumn{1}{c}{\scriptsize  fNIRS + EEG + PE}\\
  \scriptsize ELM  & \multicolumn{2}{c}{\scriptsize 54.2} & \multicolumn{1}{c}{\scriptsize 55.1} &\multicolumn{3}{c}{\scriptsize 48.2}&\multicolumn{1}{c}{\scriptsize  fNIRS + EEG + PE }\\  
\hline
%====================================================
%**************************************************
	%	[1ex] % [1ex] adds vertical space
		
		\end{tabular}
		\centering % used for centering table
		\label{table:nfm_results} % is used to refer this table in the text
\end{table}

\begin{figure*}[!t]
\centering
\includegraphics[width= 0.7 \textwidth]{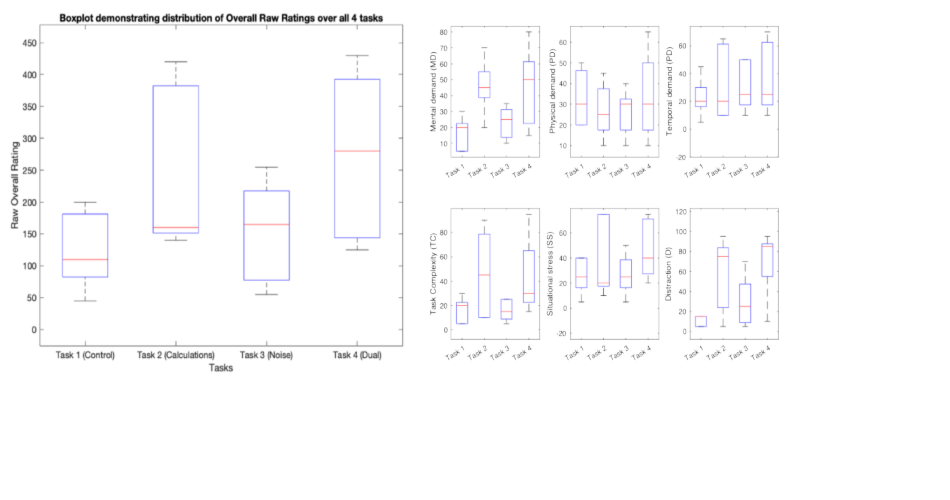}
\caption{\label{fig:boxes} Overall SURG-TLX results for each task.}
\end{figure*}
\begin{figure*}[!t]
\centering
\includegraphics[width= 0.7 \textwidth]{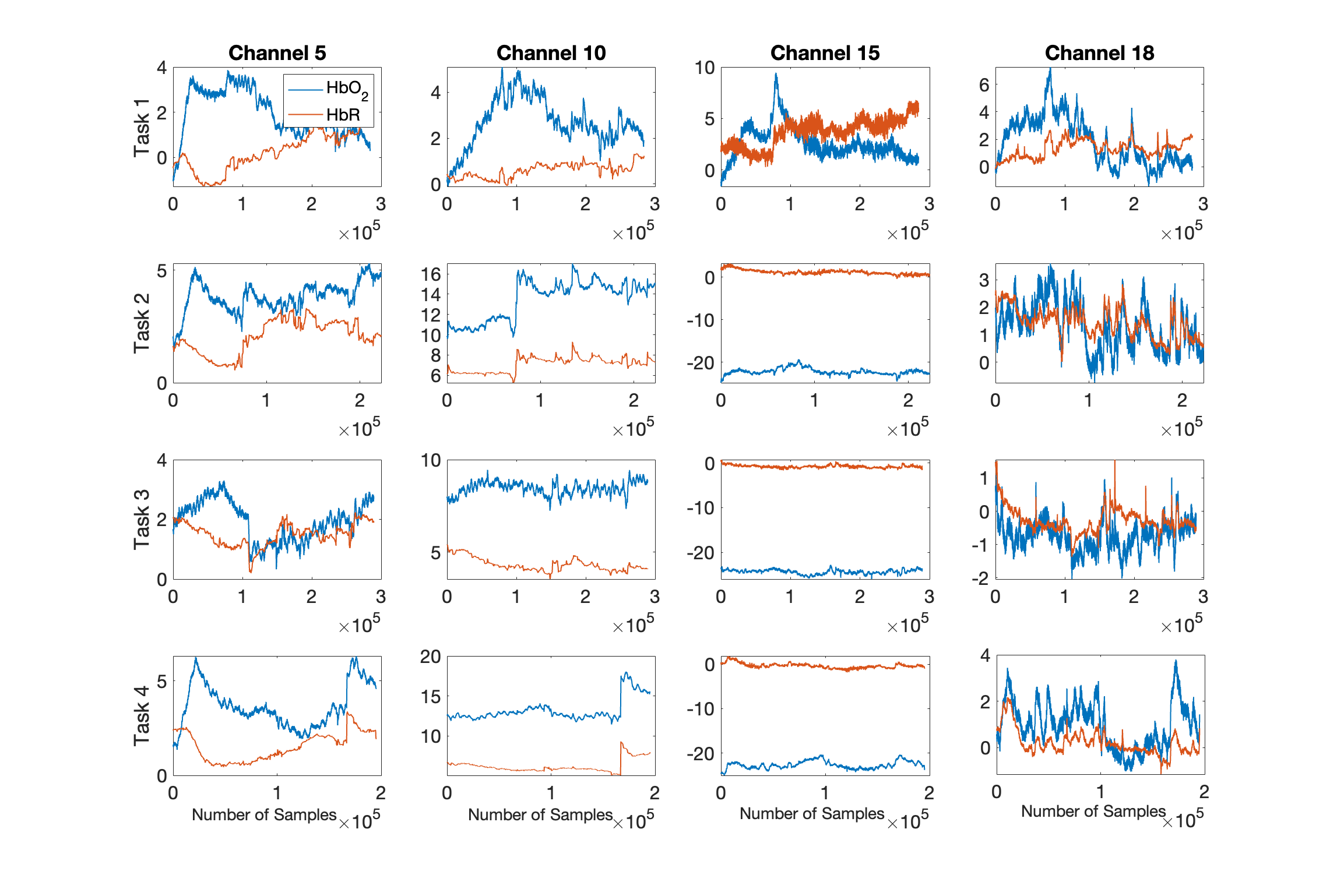}
\caption{\label{fig:channels} Change of Haemoglobin during experiments: a) HbO$_2$ and b) HbR of five interns for channels 5, 10, 15 and 18.}
\end{figure*}
\subsection{Subjective Measures: SURG-TLX }
% The SURG-TLX is an adapted and validated two-part instrument derived from the National Aeronautics and Space Task Load Index (NASA-TLX) used to measure perceived cognitive workload specifically designed to be used in surgical settings [30]. The first part consists of a multidimensional rating scale of six cognitive domains that include mental demands, physical demands, temporal demands, task complexity, situational stress and distractions. The second part requires participants to rank each domain using pairwise comparisons to provide an overall workload score.
In the surgical setting, the SURG-TLX has been used as a two-part instrument to provide subjective feedback of workload during procedures. 
 The first part consists of a multidimensional rating scale that evaluates six workload domains namely, mental demand, physical demand, temporal demand, task complexity, situational stress and distractions. The second part requires each participant to rank each given domain using pairwise comparisons to calculate the overall workload score \cite{wilson2011development}. In figure \ref{fig:boxes}, a mean score for each dimension and total CWL was is demonstrated. The perceived subjective effect of the task conditions can be observed particularly during task 2 and 4, where an increase in the mental demand is seen. This increase in the mental demand can also be observed in each cognitive dimension as illustrated in figure \ref{fig:boxes}. 

\subsection{Hemodynamic Activity}

\begin{figure}[!t]
\centering
\includegraphics[width= 0.5 \textwidth]{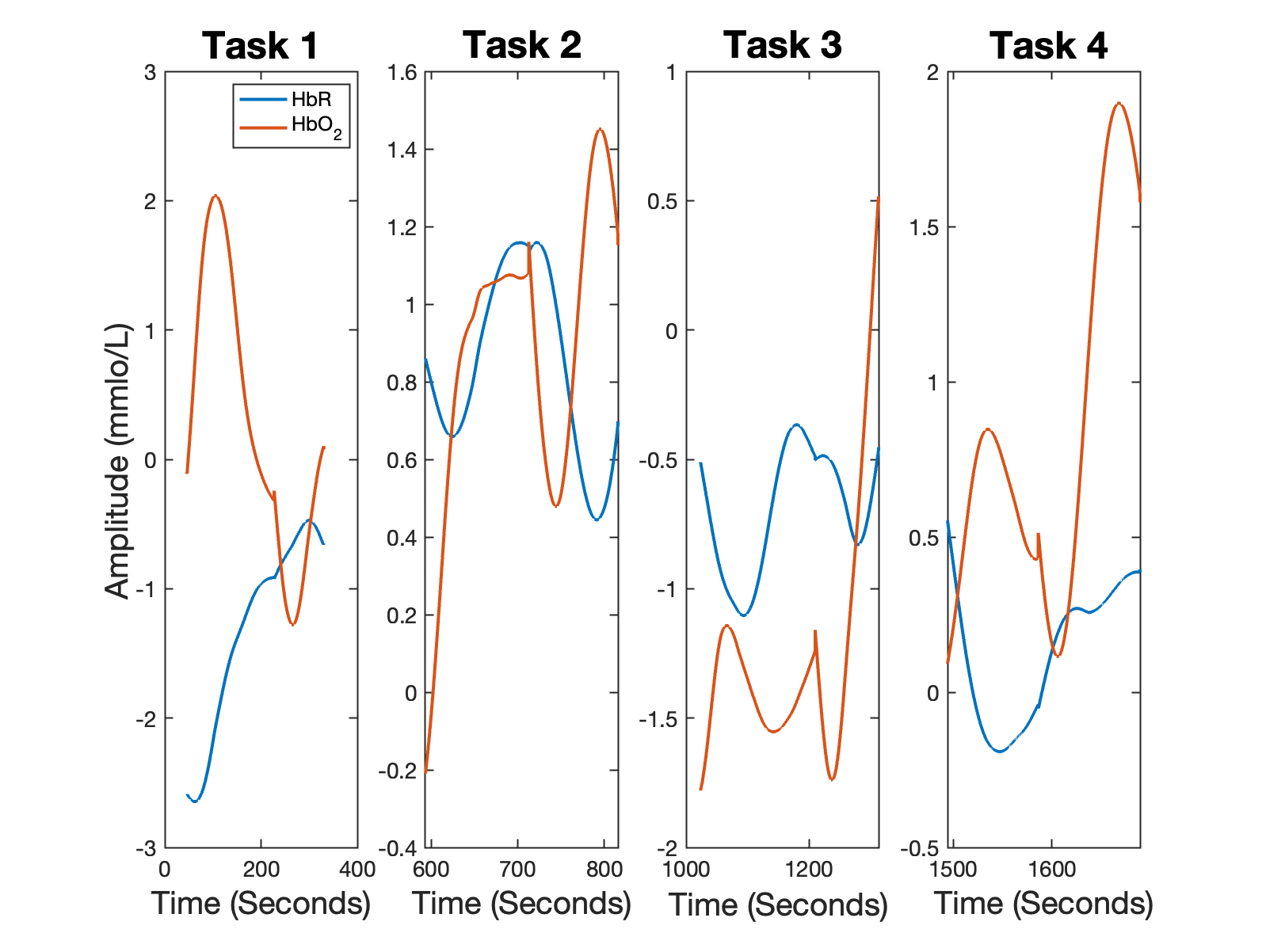}
\caption{\label{fig:haemoglobin} Average Haemoglobine activation response: a) HbO$_2$ and b) HbR of five interns.}
\end{figure}

Functional near-infrared spectroscopy (fNIRS) has been a well known non-invasive optical neuroimaging technique to monitor the brain activity. To distinguish the different levels of cognitive demand, the application of deep learning has not been extensively investigated \cite{ho2019discrimination}. Figure \ref{fig:channels} summarizes the overall changes of levels of HbO$_2$ oxyhemoglobin and HbR deoxyhemoglobin for channels 5, 10, 15 and 18 that correspond to each task. Correspondingly, in figure \ref{fig:haemoglobin}, the average of hemodynamic responses of the 18 channels during task simulation for the five subjects is presented. From the result, it can be observed the overall difference in the levels of HbO$_2$ and HbR indicating the hemodynamic changed significantly in the presence of additional distraction stressors for task 2 and 4 and then an increase for CWL of type 2 and 4 respectively. From figure \ref{fig:haemoglobin}, a significant increase in the levels of HbO$_2$ is observed at the end of task 2 and 4, in which combined distractors are considered. Wihtin this context, 

The remaining tasks exhibited a task-related decrease in HbO$_2$, in particular in the levels of HbR. From figure \ref{fig:boxes} and \ref{fig:haemoglobin}, it can be concluded that by adding a hospital bleep sounds as distractor does not add an increase in the mental demand, while substraction does. 

\section{Conclusions and Future Work}

A multimodal methodology based on the concept of transfer learning and deep learning for the identification of cognitive workload (CWL) during surgical tasks is suggested. The proposed strategy is a two-step sequential machine learning approach that first identifies if a surgeon is experiencing a mental overload in the operating theatre. At this stage, data collected from three different sensors, i.e. a) functional near infra-red spectroscopy (fNIRS), b) electroencephalography (EEG) and c) eye pupil diameter are used to feed both stages of the proposed strategy. Such data are collected by a Multi-sensing AI Environment for Surgical Task $\&$ Role Optimisation (MAESTRO) developed at the Hamlyn Centre, Imperial College London . In the first step, transfer learning is implemented by using an AlexNet Convolutional Neural Network in which its last layer is trained using its current parameters with the data collected from MAESTRO. The data is transformed into a scalogram image used to feed the AlexNet network and an Adam optimizer is applied to train its last layer (classifier). Secondly, based on this information the type of surgical task is determined using a Convolutional Neural Network in which data is coded in a 1D vector array. From our experiments, in the first step, a model accuracy of 100$\%$ was achieved by transfer learning, while in the second step an overall accuracy of 93$\%$ was produced. 

Based on our experiments, the proposed multimodal deep learning approach allows the concatenation of information coming from different sensing sources during surgical tasks while allowing complex temporal and spatial correlations with a high accuracy and low computational cost. The concept of CWL in this paper is associated with an increase mental load with surgical tasks that usually demand simultaneous processing of large amounts of information and repetitive tasks. Therefore, a multimodal deep learning strategy that allows an accurate identification of surgical tasks associated with an increase of metal work was suggested.   

Future work involves the development of a multimodal deep learning strategy that directly correlates the type of CWL with different levels of brain activity and Haemoglobin. 
\section{ Acknowledgements  }
\textbf{Funding.} This project is funded by the EPSRC Transformative Healthcare Technologies grant EP/W004755/1.
%+++++++++++++++++++++++++++++++++++++++++++++++++++++++++

%%%%%%%%% REFERENCES
{\small
\bibliographystyle{ieee_fullname}
\bibliography{egbib}
}

\end{document}